\documentclass{article}

     \PassOptionsToPackage{numbers, compress}{natbib}
 \usepackage[eandd, nonanonymous]{neurips_2026}
\nolinenumbers

\usepackage[utf8]{inputenc} % allow utf-8 input
\usepackage[T1]{fontenc}    % use 8-bit T1 fonts

\usepackage{hyperref}       % hyperlinks
\usepackage{url}            % simple URL typesetting
\usepackage{booktabs}       % professional-quality tables
\usepackage{amsfonts}       % blackboard math symbols
\usepackage{nicefrac}       % compact symbols for 1/2, etc.
\usepackage{microtype}      % microtypography
\usepackage{graphicx}
\usepackage{subcaption}
\usepackage[shortlabels]{enumitem}
\usepackage{amsmath, amssymb}
\usepackage{pifont}
\usepackage{listings}
\usepackage[dvipsnames]{xcolor}

\usepackage{enumitem}
\setlist[itemize]{leftmargin=1.2em}

\title{TailedTS: Benchmark Dataset for Heavy-Tailed Time Series Prediction and Periodicity Quantification}

\author{%
  Xinyu Chen \\
  University of Central Florida \\
  Orlando, FL 32816 \\
  \texttt{chenxy346@gmail.com} \\
  \And
  HanQin Cai \\
  University of Central Florida \\
  Orlando, FL 32816 \\
  \texttt{hqcai@ucf.edu} \\
  \AND
  Lijun Ding \\
  University of California, San Diego \\
  La Jolla, CA 92093 \\
  \texttt{l2ding@ucsd.edu} \\
  \And
  Jinhua Zhao \\
  Massachusetts Institute of Technology \\
  Cambridge, MA 02139 \\
  \texttt{jinhua@mit.edu} \\
}

\begin{document}

\maketitle

\begin{abstract}

We present TailedTS, a large-scale benchmark dataset derived from Wikipedia hourly page view observations throughout 2024, specifically designed to test time series forecasting models under heavy-tailed, zero-inflated, and non-Gaussian conditions. The dataset comprises approximately 24.69 billion data points spanning roughly 3 million unique Wikipedia pages per month, stored in high-efficiency Apache Parquet format. Wikipedia traffic follows a pronounced power-law distribution where roughly 5\% of pages account for over 70\% of total page views, creating a natural and rigorous testbed for model robustness against extreme volatility that are absent from or underrepresented in existing benchmarks such as M4, M5, and UCI electricity datasets. TailedTS enables several research tasks. First, we introduce a periodicity quantification framework based on sparse autoregression with sparsity and non-negativity constraints, revealing that frequently-viewed pages exhibit significantly weaker periodic structure than their less-viewed counterparts, showing direct implications for server allocation and traffic forecasting on large digital platforms. Second, we provide standardized prediction benchmarks evaluated under a suite of non-Gaussian loss functions, including $\ell_1$-norm, Huber, quantile, and $\ell_p$-norm losses, demonstrating that standard Gaussian-based estimators degrade substantially on high-volume page categories, while robust alternatives provide consistent gains across all traffic scales. TailedTS is publicly available at \url{https://doi.org/10.5281/zenodo.17070469}.

\end{abstract}

\section{Introduction}

Heavy-tailed data are ubiquitous in real-world applications, spanning fields from statistics to machine learning \cite{nair2022fundamentals, clemenccon2025weak, gurbuzbalaban2021heavy, roy2021empirical}. In time series analysis, identifying and predicting patterns in such data is critical, yet challenging. Traditional predictive models often rely on Gaussian assumptions for residuals, which are frequently violated by the heavy-tailed behavior found in empirical datasets. While existing benchmarks such as the M-Competitions \cite{makridakis2020m4, makridakis2022m5} or UCI electricity datasets \cite{electricityloaddiagrams20112014_321} capture various temporal dynamics, they often struggle to represent the extreme, event-driven volatility and high-dimensional sparsity inherent in complex human interest patterns.

In this work, we introduce \textbf{TailedTS}, a benchmark dataset for heavy-tailed time series prediction and periodicity quantification. Based on Wikipedia page view observations, this dataset exhibits a clear power-law distribution where a small subset of trending pages captures the majority of global attention. 
% From our preliminary analysis, the 3 million popular Wikipedia pages have gained around 70\% page views among all 60 million pages on Wikipedia. 
As the power-law distribution of Wikipedia page view time series is quite remarkable, it demands us to find appropriate machine learning models for analyzing and predicting these heavy-tailed time series.
% \ld{What does power law of the data mean here? Is that the 3 million pages captures 70\% views part of the power law or what? If not, what is the purpose of mentioning it? If yes, why it is part of the power law?}
In this work, our contributions are three-fold:
\begin{itemize}
\item \textbf{Benchmark Dataset}: We provide a high-dimensional, heavy-tailed, and zero-inflated time series dataset that allows the research community to evaluate machine learning models against non-Gaussian residuals and stochastic bursts. The TailedTS dataset include millions of time series of a long time period (i.e., 8784 hourly time steps throughout 2024), involving complicated periodic patterns and extrema. While most time series models stem from Gaussian residual assumption and violate the heavy-tailed behavior, the TailedTS dataset allows the research community to examine time series modeling tasks in the case of heavy-tailed noises.
\item \textbf{Periodicity Quantification}: We utilize sparse autoregression with sparsity and non-negativity constraints to automatically identify dominant auto-correlations, such as daily and weekly cycles. Our analysis reveals that page view time series for popular Wikipedia pages are notably less periodic than their less-viewed counterparts. On the digital platform, while the popular Wikipedia pages are the heavy tails of the whole dataset, predicting their future page views is challenging. 
% optimally allocating cloud servers
% \ld{Is the last sentence also part of the contribution?}
\item \textbf{Robust Predictive Modeling}: We evaluate autoregressive prediction models by using a variety of non-Gaussian loss functions, including Huber loss, quantile loss, and $\ell_p$-norm losses, to improve resilience against heavy-tailed noise and volatility spikes. 
% \ld{Is the loss part of the predictive model?} 
Although there are several ways to improve the performance of time series prediction with machine learning models, we examine the choice of loss functions and verify the importance of modeling non-Gaussian residuals with these loss functions on heavy-tailed time series. By integrating TailedTS with existing meteorological benchmarks, researchers can better assess the scalability of modern frameworks and the efficacy of robust loss functions in capturing irregular, interconnected human interest patterns.
\end{itemize}

% Reformulating the predictive models in following directions: i) paradigm, ii) objective function (e.g., loss function over residuals and penalty terms), iii) constraints.

\section{Related Work}

\subsection{Existing Datasets}

\begin{figure}[ht!]
\centering
\begin{subfigure}[b]{0.24\textwidth}
\centering
\includegraphics[height=2.65cm]{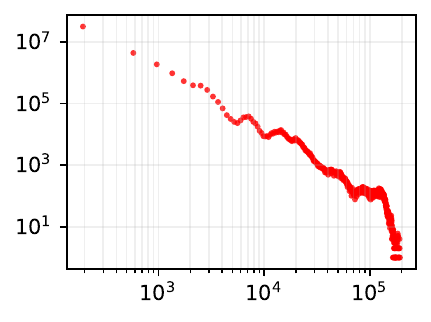}
\caption{UCI electricity data \cite{electricityloaddiagrams20112014_321}}
\end{subfigure}
\begin{subfigure}[b]{0.24\textwidth}
\centering
\includegraphics[height=2.65cm]{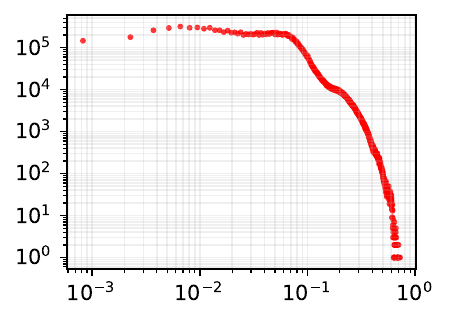}
\caption{SF traffic data \cite{lai2018modeling}}
\end{subfigure}\hspace{0.3em}
\begin{subfigure}[b]{0.24\textwidth}
\centering
\includegraphics[height=2.65cm]{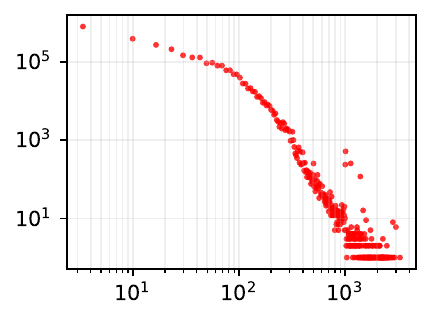}
\caption{Air quality data \cite{godahewa_2020_4656756}}
\end{subfigure}
\begin{subfigure}[b]{0.24\textwidth}
\centering
\includegraphics[height=2.65cm]{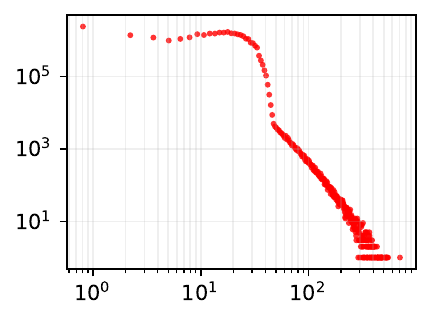}
\caption{Weather data \cite{godahewa_2020_4654822}}
\end{subfigure}
\caption{Log-log histograms of time series datasets. The plots display the frequency distribution of (a) UCI electricity load diagrams, (b) SF traffic occupancy, (c) Air quality indices, and (d) Meteorological weather variables on a logarithmic scale for both axes. The approximately linear decay observed in several of these plots is a characteristic indicator of heavy-tailed distributions. 
% These visualizations demonstrate that extreme values or outliers occur with significantly higher frequency than would be expected in a Gaussian distribution, posing a distinct challenge for traditional prediction models.
}
\label{data_histogram_log}
\end{figure}

The evaluation of time series prediction or forecasting models relies heavily on a diverse set of benchmark datasets that capture various temporal dynamics, ranging from seasonal patterns to stochastic volatility. These datasets exhibit a wide range of empirical distributions and highlight the non-Gaussian nature and varying scales of real-world time series data, often characterized by significant skewness and heavy-tailed behavior, see Figures~\ref{data_histogram_log} and \ref{data_histogram}. However, these datasets are not power-law distributed.

In the domain of retail and economics, the M-Competitions (e.g., M4 \cite{makridakis2020m4} and M5 \cite{makridakis2022m5}) remain the most influential benchmarks. The M4 dataset provides a massive collection of 100,000 univariate series across multiple frequencies \cite{makridakis2020m4}, serving as a rigorous test for model generalizability. The M5 retail data, sourced from Walmart, introduces hierarchical complexities and exogenous variables such as price and promotions \cite{makridakis2022m5}, shifting the focus toward intermittent demand forecasting and the integration of static metadata. Energy consumption and infrastructure monitoring represent another critical pillar of time series research. The UCI electricity load diagrams and individual household power consumption datasets are frequently used to evaluate models on high-frequency, multivariate data \cite{hebrail2012individual, electricityloaddiagrams20112014_321}. These datasets often display strong diurnal and weekly periodicity, but as shown in our log-log analysis (see Figure~\ref{data_histogram_log}), they also contain extreme consumption events that deviate from Gaussian assumptions. Similarly, the SF traffic occupancy and solar energy datasets provide spatial-temporal challenges \cite{lai2018modeling}, where the objective is to capture the correlation between hundreds of sensors or power plants over long forecasting horizons.

% \ld{So previous data is non-gaussian and our dataset is also non-gaussian? What is the difference? I did not see the non-gaussian part or heavy tail part as a difference in the following paragraph.}
In this work, the power-law distributed Wikipedia page view dataset significantly complements traditional benchmarks such as M5 retail or UCI electricity by introducing a distinct scale of high-dimensional sparsity and extreme, event-driven volatility that deviates from the well-defined periodicity. As evidenced by the log-log power-law distribution in Figure~\ref{wikipedia_page_dist}(a), Wikipedia traffic is governed by a long-tail dynamic where a minute fraction of trending pages captures the vast majority of global attention, providing a rigorous stress test for a model's robustness against non-Gaussian outliers and stochastic bursts. By integrating this dataset alongside existing meteorological benchmarks, researchers can better evaluate the scalability of GPU-native frameworks and the efficacy of robust loss functions such as the Huber loss in capturing complex, interconnected human interest patterns that are far more irregular than the physical constraints found in transportation or climate systems.

\subsection{Literature Review}
%We review related work in two main directions: Classical and modern time series prediction methods, and robust autoregressive modeling under heavy-tailed noise.

To contextualize the evaluation challenges posed by heavy-tailed time series analysis, we review representative modeling approaches for time series prediction and robust autoregressive estimation. Rather than providing an exhaustive survey, we focus on the modeling assumptions and loss functions that implicitly shape benchmark design and evaluation practices.%, highlighting methodological gaps that existing datasets often fail to expose.

\subsubsection{Time Series Prediction}

Time series prediction is a core problem in machine learning, with applications spanning economics, environmental science, engineering, and the natural sciences. Classical approaches emphasize parametric models that capture temporal dependence through linear dynamics, such as autoregressive (AR), moving average (MA), and ARIMA models. Under stationarity and light-tailed noise assumptions, these models admit efficient estimation and strong theoretical guarantees \cite{brockwell2009time,box2015time}.

Recent advances in machine learning have expanded the modeling landscape. RNNs and their gated variants, such as LSTM and GRU, enable flexible modeling of nonlinear temporal dependencies \cite{hochreiter1997long,chung2014empirical,cho2014learning}. 
Temporal convolutional networks provide an alternative based on causal convolutions and dilation \cite{lea2017temporal,bai2018empirical}.
Attention-based models and Transformer variants further improve long-range dependency modeling and achieve strong empirical performance on large-scale benchmarks \cite{li2019enhancing,zhou2021informer,wu2021autoformer}. More recently, LLMs have been reprogrammed for time series forecasting by converting numerical sequences into prompt-based representations, enabling zero-shot and few-shot prediction in a range of benchmark settings \cite{jin2023time,gruver2023large,liu2024autotimes}. Despite the diversity of models, time series predictors, from classical methods to deep learning and LLM based approaches, are typically evaluated using mean squared error or likelihood-based objectives that emphasize average case performance and implicitly favor well-behaved noise distributions. 
%Consequently, standard benchmarks often overlook behavior under heavy-tailed noise and provide limited assessment of a model’s ability to recover periodic structure.

%Time LLM \cite{jin2023time}

% Time series prediction has a long history rooted in stochastic process modeling, with autoregressive (AR) \cite{???}, moving average (MA) \cite{???}, and the combined ARMA models \cite{???} serving as foundational approaches for linear temporal dependence. These classical models assume stationarity and rely heavily on second-order statistics, making them effective under Gaussian or light-tailed noise but potentially fragile in the presence of heavy-tailed or impulsive observations.

% Beyond linear models, spectral methods \cite{???} provide an alternative viewpoint by characterizing temporal structure in the frequency domain. Periodicity detection and forecasting can be performed via Fourier-based representations, which are closely connected to AR processes through their spectral formulation. However, standard spectral estimators also rely on moment conditions and are sensitive to outliers, limiting their effectiveness under heavy-tailed regimes \cite{???}.

\subsubsection{Robust Autoregressive Models} 
AR models are one of the most fundamental tools for time series analysis, but classical AR procedures typically rely on Gaussian assumptions and finite second-order moments, making them fragile under heavy-tailed or impulsive noise. When innovations follow heavy-tailed distributions, the variance may be infinite, invalidating Least Squares (LS) (i.e., $\ell_2$-norm) based inference \cite{samorodnitsky1994stable}. This has motivated a long line of work on robust alternatives based on loss functions
%(i.e., M-estimators) 
with reduced sensitivity to outliers.

%Early work addressed this limitation through M-estimators \cite{huber1992robust}, which replace quadratic loss with sub-quadratic alternatives to mitigate the influence of outliers. Extensions to time series models include robust AR estimations via \textit{least absolute deviations} \cite{davis1997least} and bounded-influence estimators \cite{maronna2019robust}.

%\vspace{0.05in}
\noindent\textbf{Least Absolute Deviations ($\ell_1$-norm Loss).}
Least absolute deviations (LAD) estimation replaces the quadratic loss with an $\ell_1$-norm based objective, yielding estimators with bounded influence and improved robustness under heavy-tailed noise. Early work established consistency and asymptotic normality of LAD estimators for AR models under mild conditions \cite{davis1997least}. More recent analyses show that LAD estimators remain well-behaved under heavy-tailed and conditional heteroskedastic noise, with convergence rates depending explicitly on the tail index \cite{zhang2022lade}. Numerous extensions of LAD have been developed for various AR settings \cite{breidt2001least,ling2005self,xia2008generalized,wu2010least,liu2020least}.

%\vspace{0.05in}
\noindent\textbf{Huber Loss.}
Huber loss interpolates between $\ell_2$- and $\ell_1$-losses by applying quadratic penalization to small residuals and linear penalization to large residuals. As a result, Huber loss retains statistical efficiency under light-tailed noise while limiting the influence of outliers \cite{huber1992robust}. In time series settings, Huber-type estimators have been extended to AR models, offering a favorable bias--variance trade-off compared to pure LAD or least squares \cite{maronna2019robust}. Recent work studies adaptive selection of the Huber threshold under heavy-tailed distributions, showing that properly tuned estimators achieve near-optimal statistical rates under weak moment assumptions \cite{sun2020adaptive,fan2022adaptive}.

%\vspace{0.05in}
\noindent\textbf{Quantile Losses.}
Quantile loss, a.k.a.~pinball loss, enables the estimation of conditional quantiles rather than the conditional mean in regressions by introducing asymmetric linear penalties \cite{koenker2001quantile}. This framework naturally generalizes LAD and provides robustness to heavy-tailed noise while capturing distributional asymmetries. It has been extended to AR models and characterizes the full conditional distribution beyond mean dynamics \cite{koenker2006quantile}. Recent work applies quantile AR models to capture heterogeneous and distributional dynamics in temporally and spatially dependent data, such as environmental and economic time series \cite{hadi2025spatial,chavas2020quantile,castillo2023spatial}.

%\vspace{0.05in}
\noindent\textbf{$\ell_p$-norm Loss.}
Losses based on $\ell_p$-norms with $p \in (0,1)$ provide a more aggressive form of robustness by penalizing residuals sublinearly. Compared to $\ell_1$-norm loss, these objectives further downweight large deviations, making them particularly suitable for heavy-tailed or impulsive noise where extreme observations dominate the sample. 
However, these losses are nonconvex and non-smooth, leading to challenging optimization landscapes with potential local minima. Despite the difficulties, iterative reweighting and majorization--minimization approaches have been developed to compute tractable approximate solutions \cite{o1990robust,daubechies2010iteratively,peng2022global,sun2016majorization}. In the AR context, $\ell_p$-norm losses have received comparatively limited attention and offer a potential nonconvex alternative in heavy-tailed regimes.

\section{Data Description}

\begin{figure}[ht!]
\centering
\begin{subfigure}[b]{0.28\textwidth}
\centering
\includegraphics[width=\textwidth]{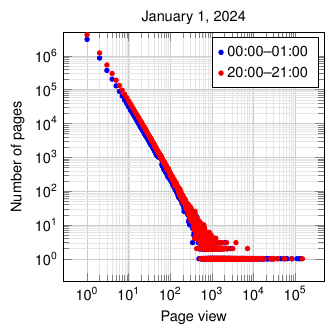}
\caption{Log-log plot}
\end{subfigure}\hspace{1em}
\begin{subfigure}[b]{0.37\textwidth}
\centering
\includegraphics[width=\textwidth]{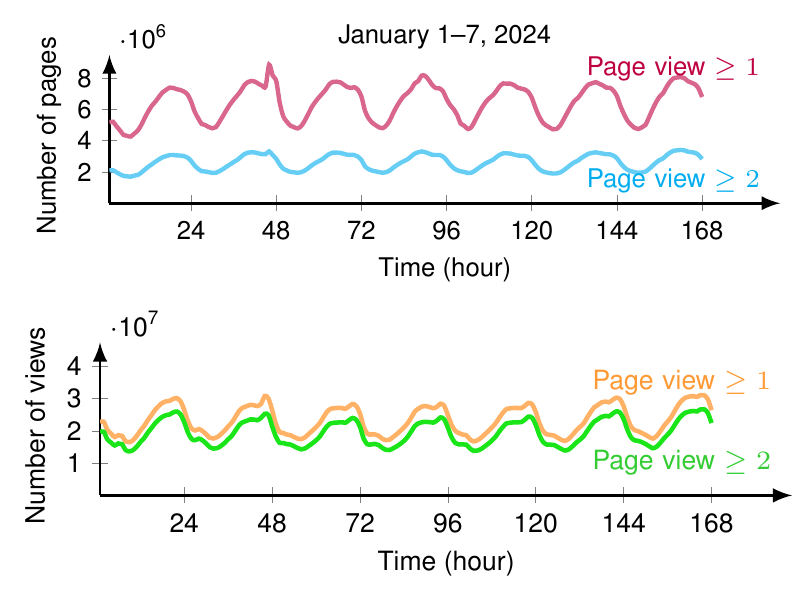}
\caption{Aggregated hourly time series}
\end{subfigure}\hspace{1em}
\begin{subfigure}[b]{0.27\textwidth}
\centering
\includegraphics[width=\textwidth]{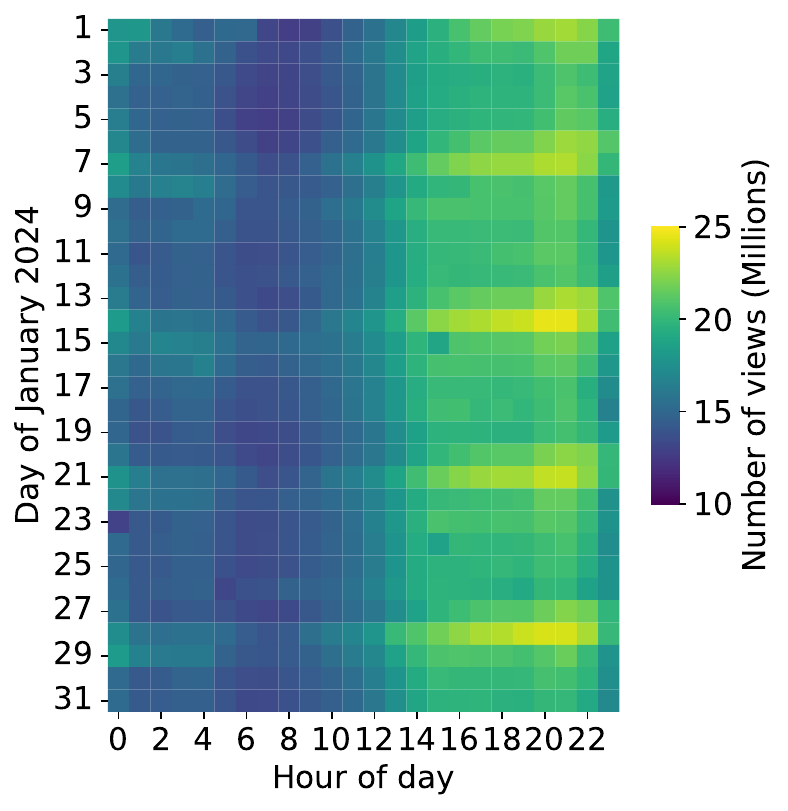}
\caption{Aggregated time series}
\end{subfigure}
\caption{Empirical demonstration of the Wikipedia page view time series dataset for January 2024. (a) A log-log plot of the number of pages versus page views, indicating a clear power-law distribution. (b) Aggregated hourly time series over the first week of January, showcasing strong 24-hour seasonality and diurnal cycles. (c) A heatmap of aggregated time series by hour of day across the entire month, highlighting consistent peak activity periods (yellow/green) and late-night troughs (blue/purple), reflective of global user engagement patterns.}
\label{wikipedia_page_dist}
\end{figure}

\textbf{Data Patterns.} The Wikipedia page view data portal includes long-term hourly page view observations since May 2015 across more than 60 million Wikipedia pages.\footnote{\url{https://dumps.wikimedia.org/other/pageviews/readme.html} under the Creative Commons CC0 dedication.} As shown in Figure~\ref{wikipedia_page_dist}(a), the empirical distribution of Wikipedia page views that uses logarithmic scales on both the horizontal and vertical axes at two certain hours on January 1, 2024 demonstrates heavy tails. That means that only a small fraction of Wikipedia pages have been frequently-viewed, demonstrating power-law distributions.
% \ld{If it is gaussian, what would the plot look like? How do you deal with negative numbers?} 
By observing the time series of the number of pages and page views from January 1, 2024 to January 7, 2024 in the top panel of Figure~\ref{wikipedia_page_dist}(b), they demonstrate strong periodic patterns across different days. Although the pages whose number of page views greater than 1 (i.e., page view $\geq2$) are only a small portion of total Wikipedia pages, their page views are dominant as shown in the bottom panel of Figure~\ref{wikipedia_page_dist}(b). Thus, it motivates us to pre-process the page view data and extract the frequently-viewed pages. One remarkable advantage by doing so is mitigating the impact of zero-inflated and biased observations when learning a certain machine learning model.

% \ld{May want to give a paragraph title of each paragraph in this section.}
\textbf{Data Alignment.} To get the time series from the page view observations, the first task is data alignment across different Wikipedia pages. In the original page view observations, the observed Wikipedia pages which have been viewed at least once are quite different between two hours. For example, Figure~\ref{wikipedia_page_dist}(b) shows that the total number of pages are changing over time, presenting the predictable ebbs and flows of web traffic data. The number of pages in peak hours is significantly greater than off-peak hours. By selecting the Wikipedia pages that have been viewed at least 10 times in each day, we can roughly obtain 5 million unique pages. We build the sample time series of hourly page views for each day. Of the source data in January 2024, we identify the intersection set of pages across 31 days, referring to 3,031,046 unique Wikipedia pages. Although the selected 3 million pages of January 2024 are less than 5\% in total Wikipedia pages, they have 12.88 billion total views and dominate the total page views in 72\% across the whole month of January 2024. Figure~\ref{wikipedia_page_dist}(c) shows the total number of page views of each hour in January 2024, demonstrating a remarkable weekly periodicity where weekends usually have higher page views during peak hours than weekdays.

\textbf{Dataset Preparation.} We provide the long-term time series dataset throughout 2024, exceeding 24 billion data points, see Table~\ref{data_file_table} in Supplementary Material. To ensure data quality and focus on frequently viewed Wikipedia pages, each monthly subset includes only those pages that maintained a minimum threshold of 10 views per day, every day for that month. Accordingly, the proportion of zero page views in this dataset is around 24\%. This filtering results in a robust dataset covering approximately 3 million unique Wikipedia pages and over 2 billion discrete data points per month.

\section{Problem Formulation}

\subsection{Time Series Periodicity Quantification}

Periodicity is a key time measurement of sequences, and a clear periodicity pattern often implies strong predictability when using time series models. In this work, we quantify the time series periodicity by identifying the dominant non-negative auto-correlations from autoregressive process. Since the multivariate time series of Wikipedia page views can be roughly classified as different categories according to the topic or the popularity of Wikipedia pages, we introduce several categories indexed by $\gamma\in[G]$, and each category includes a brunch of time series $\{x_{i,\gamma,t}\}_{i,t}$. The optimization problem of sparse autoregression can be formulated as follows,
\begin{equation}\label{sparse_autoregression_opt}
\begin{aligned}
\min_{\{\boldsymbol{w}_{\gamma}\in\mathbb{R}_{+}^{d}\}_{\gamma\in[G]}}\quad&\sum_{\gamma\in[G]}\sum_{i\in[n_\gamma]}\Bigl(x_{i,\gamma,t}-\sum_{k\in[d]}w_{\gamma,k}x_{i,\gamma,t-k}\Bigr)^2 \\
\text{s.t.}\quad&\|\boldsymbol{w}_{\gamma}\|_0\leq\tau\in\mathbb{Z}^{+},\quad\operatorname{supp}(\boldsymbol{w}_h)=\operatorname{supp}(\boldsymbol{w}_\gamma),\quad\forall h,\gamma\in[G],\,h\neq\gamma, \\
\end{aligned}
\end{equation}
where the upper bound of $\ell_0$-norm (denoted by $\|\cdot\|_0$) of coefficient vector $\boldsymbol{w}_{\gamma}\in\mathbb{R}^{d}$ is sparsity level $\tau$, which is far smaller than the order $d\in\mathbb{Z}^{+}$. In the model, the support sets (denoted by $\operatorname{supp}(\cdot)$) of $\boldsymbol{w}_{\gamma},\gamma\in[G]$ across different categories are assumed to consistent.

% The sparse autoregression is an efficient interpretable machine learning method for quantifying periodic patterns of real-world time series data with inherent periodic cycles. 
On the top of time series autoregression, the sparse autoregression assumes both sparsity and non-negativity of auto-correlations for automatically identifying dominant auto-correlations such as daily and weekly cycles as interpretable results \cite{chen2025interpretable}. For each category $\gamma\in[G]$, the model only uses a sparse coefficient vector $\boldsymbol{w}_{\gamma}$ to express the dominant auto-correlations in a sequence of time series. Thus, the seasonality of each group represents the overall periodic pattern of thousands or millions of page view time series. In theory, the optimization problem in Eq.~\eqref{sparse_autoregression_opt} can be seamlessly reformulated as a Mixed-Integer Quadratic Programming (MIQP) such that
\begin{equation}\label{sparse_autoregression_opt_quad}
\begin{aligned}
\min_{\{\boldsymbol{w}_{\gamma}\}_{\gamma\in[G]},\,\boldsymbol{\beta}}\quad&\sum_{\gamma\in[G]}\bigl(\operatorname{tr}(\boldsymbol{w}_{\gamma}\boldsymbol{w}_{\gamma}^\top\boldsymbol{\Phi}_{\gamma})-2\boldsymbol{w}_{\gamma}^\top\boldsymbol{\psi}_{\gamma}\bigr) \\
\text{s.t.}\quad&0\leq\boldsymbol{w}_{\gamma}\leq\mathcal{M}\cdot\boldsymbol{\beta},\quad\forall\gamma\in[G],\quad\|\boldsymbol{\beta}\|_1\leq\tau\in\mathbb{Z}^{+},\quad\boldsymbol{\beta}\in\{0,1\}^{d},
\end{aligned}
\end{equation}
where the operator $\operatorname{tr}(\cdot)$ denotes matrix trace. The binary decision vector is $\boldsymbol{\beta}\in\{0,1\}^{d}$ with entries that selected from either 0 and 1, while $\mathcal{M}\in\mathbb{R}_{+}$ is a sufficiently large constant. The decision variables include both coefficient vectors $\boldsymbol{w}_\gamma,\gamma\in[G]$ and binary vector $\boldsymbol{\beta}\in\mathbb{R}^{d}$. In particular, we simplify the objective in Eq.~\eqref{sparse_autoregression_opt} as the current formula in which the data pair is defined as follows,
\begin{equation}
\boldsymbol{\Phi}_{\gamma}=\sum_{i\in[n_{\gamma}]}\boldsymbol{A}_{i,\gamma}^\top\boldsymbol{A}_{i,\gamma}\in\mathbb{R}^{d\times d},\quad\boldsymbol{\psi}_{\gamma}=\sum_{i\in[n_\gamma]}\boldsymbol{A}_{i,\gamma}^\top\boldsymbol{y}_{i,\gamma}^{(0)}\in\mathbb{R}^{d},\quad\forall\gamma\in[G],
\end{equation}
where $\boldsymbol{y}_{i,\gamma}^{(j)}=\{x_{i,\gamma,t}\}_{t\in[d+1-j,T-j]}\in\mathbb{R}^{T-d},\,\forall j\in[0,d]$ and $\boldsymbol{A}_{i,\gamma}=\{\boldsymbol{y}_{i,\gamma}^{(k)}\}_{k\in[d]}\in\mathbb{R}^{(T-d)\times d}$, which are constructed from the time series data.

\subsection{Heavy-Tailed Time Series Prediction}

% \ld{Predictor $f$ as a function of $\mathbb{R}^d $ to $\mathbb{R}$ is a bit weird. Also, did we use complicated $f$ other than the AR(d) model?}
On the time series data such as $\boldsymbol{x}\in\mathbb{R}^{T}$, the predictor model $f:\mathbb{R}^{d}\to\mathbb{R}$ can be used to learn auto-correlations from these data points. The residual at time $t$ is written as $\varepsilon_{t}=x_t-f(\boldsymbol{w};\{x_{t-k}\}_{k\in[d]})$ where $d$ is the order of predictor, and the coefficient vector $\boldsymbol{w}\in\mathbb{R}^{d}$ represents coefficients. In the case of Gaussian residuals, the autoregression can be seamlessly modeled as an LS problem. However, the challenge arises as the residuals deviate from the Gaussian distribution, which is a great concern in real-world time series data. Thus, we formulate the objective function of predictive model with a general form of loss functions, i.e., the sum of $\rho(\varepsilon_t)$. The optimization problem is written as follows,
\begin{equation}
\begin{aligned}
\min_{\boldsymbol{w}}\quad \sum_{t\in[d+1,T]}\rho(\varepsilon_t)\quad\quad\text{s.t.}\quad \boldsymbol{w}\in\mathcal{C},\quad\varepsilon_t=x_{t}-f(\boldsymbol{w};\{x_{t-k}\}_{k\in[d]}),\quad\forall t\in[d+1,T],
\end{aligned}
\end{equation}
where the constraints include a feasible set $\mathcal{C}$ on the model parameters $\boldsymbol{w}$, e.g., sparsity constraints \cite{chen2025interpretable}. The optimization over $\boldsymbol{w}$ varies if we have different forms of loss function, see e.g., Table~\ref{loss_func_table} for reference. If we learn a model with certain loss functions, then the one-step ahead prediction is  $\hat{x}_{t+1}=f(\boldsymbol{w};\{x_{t+1-k}\}_{k\in[d]})$ with order $d$.

\begin{table}[h]
    \caption{Mathematical properties and optimization strategies for different residual-based loss functions in the predictive model.}
    \centering
    \begin{tabular}{c|cccc}
    \toprule
    Loss function & Formula of $\rho(\varepsilon_t)$ & Hyperparameter & Optimizer & Convexity \\
    \midrule
    $\ell_2$-norm loss & $\varepsilon_t^2$ & - & LS & \ding{51} \\
    $\ell_1$-norm loss & $|\varepsilon_t|$ & - & ISTA \cite{daubechies2004iterative, beck2009fast} \& LP \cite{boyd2004convex} & \ding{51} \\
    Huber loss & Eq.~\eqref{huber_quantile} & Huber threshold $\delta$ & QP \cite{boyd2004convex} \& IRLS \cite{o1990robust, daubechies2010iteratively} & \ding{51} \\
    Quantile loss & Eq.~\eqref{huber_quantile} & quantile $\tau\in(0,1)$ & LP \cite{koenker2001quantile} & \ding{51} \\
    $\ell_p$-norm loss & $|\varepsilon_{t}|^p$ & power $p\in(0,1)$ & IRLS \cite{daubechies2010iteratively} & \ding{55} \\
    \bottomrule
    \end{tabular}    \label{loss_func_table}
\end{table}

In Table~\ref{loss_func_table}, we summarize some non-Gaussian loss functions on residuals, including $\ell_1$-norm loss, Huber loss, quantile loss, and $\ell_p$-norm loss. Huber loss is quadratic (i.e., squared $\ell_2$-norm) for small residuals and linear (i.e., $\ell_1$-norm) for large residuals, providing robustness against outliers.\footnote{In Table~\ref{loss_func_table}, we simplify several notions such as Iterative Shrinkage-Thresholding Algorithm (ISTA), Quadratic Programming (QP), Iteratively Reweighted Least Squares (IRLS), and Linear Programming (LP).} For any residual $\varepsilon_t\in\mathbb{R}$, the Huber loss function and quantile loss function are defined as follows,
\begin{equation}\label{huber_quantile}
\rho_{\delta}(\varepsilon_t)=\begin{cases}
\varepsilon_t^2,&\text{if $|\varepsilon_t|\leq\delta$}, \\
\delta(2|\varepsilon_t|-\delta),&\text{otherwise},
\end{cases}\quad\text{and}\quad
\rho_{\tau}(\varepsilon_t)=\begin{cases}
\tau\varepsilon_t,&\text{if $\varepsilon_t\geq0$}, \\
(\tau-1)\varepsilon_t,&\text{otherwise},
\end{cases}
\end{equation}
respectively. The optimization over Huber loss is reformulated as QP and IRLS, see \ref{huber_opt_section} for details. While QP problem has an exact solution via the use of Second-Order Cone Programming (SOCP) solvers, IRLS is an efficient method for large-scale data. The optimization over quantile loss is converted into an LP problem in which the special case of quantile $\tau=1/2$ leads to an $\ell_1$-norm loss, see \ref{quantile_opt_section} for details.

\begin{figure}[ht!]
\centering
\includegraphics[width=1\linewidth]{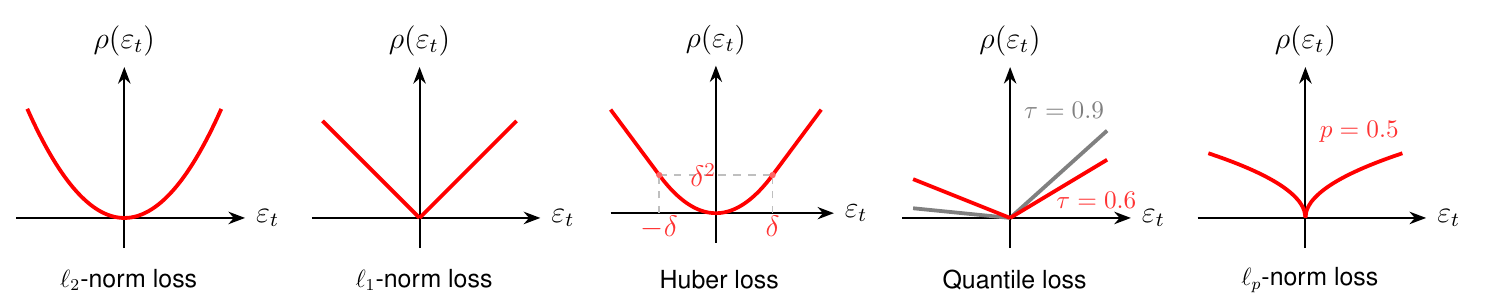}
\caption{Visual comparison of loss functions $\rho(\varepsilon_t)$ highlighting their penalty behaviors. From left to right: the symmetric $\ell_2$-norm and $\ell_1$-norm losses; the Huber loss showing the quadratic-to-linear transition at threshold $\delta$; the asymmetric quantile loss for different $\tau$ values; and the nonconvex $\ell_p$-norm loss ($0<p < 1$) demonstrating redescending-like characteristics for large residuals.}
\label{loss}
\end{figure}

When leveraging the autoregressive predictor $x_t = \sum_{k \in [d]} w_k x_{t-k} + \varepsilon_t$ to formulate the optimization problem, the choice of the objective function $\sum_t \rho(\varepsilon_t)$ fundamentally dictates the model's resilience to heavy-tailed noise distributions. In contrast to the standard $\ell_2$-norm loss, which assumes Gaussian residuals and disproportionately penalizes large errors, leading to biased parameter estimates in the presence of volatility spikes, robust losses such as the Huber and $\ell_1$ functions mitigate the influence of extreme values or outliers by capping the influence of large residuals. Specifically, the linear tails of these functions prevent the pull of outliers from dominating the gradient during optimization. Furthermore, for highly non-Gaussian environments, the nonconvex $\ell_p$-norm ($0 < p < 1$) offers a redescending-like behavior that can effectively ignore extrema, though it necessitates more sophisticated Iteratively Reweighted Least Squares (IRLS \cite{o1990robust, daubechies2010iteratively,peng2022global}) or proximal solvers to navigate the nonconvex landscape and reach a stable local minimum.

\section{Results on Wikipedia Dataset}
\label{sec:numerical results}
% In this section, we first quantify the time series periodicity of Wikipedia page views with sparse autoregression and then summarize them into three categories. Throughout preliminary seasonality quantification, we perform time series prediction and highlight the challenges of modeling heavy-tailed noises.

\subsection{Seasonality Exploration}

% \begin{figure}[ht!]
% \centering
% \subfigure[Dominant auto-correlations]{
% \centering
% \includegraphics[scale=0.9]{graphics/seasonality1.pdf}
% }\hspace{2em}
% \subfigure[Log-log plot]{
% \centering
% \includegraphics[scale=0.9]{graphics/seasonality2.pdf}
% }
% \caption{Seasonality analysis of Wikipedia page view time series with the sparse autoregression model. (a) Auto-correlations on the optimized index set that controlled by the sparsity levels $\tau=8,10$. (b) Log-log plot of distributions of page view data. The top panel shows 9 sub-categories with total number of page views $\mathcal{O}(10^3)$ in January 2024. The bottom panel shows 9 sub-categories with total number of page views $\mathcal{O}(10^4)$.}
% \label{seasonality}
% \end{figure}

% \ld{Please give a paragraph title for each paragraph.}

\begin{figure}[ht!]
\centering
    \begin{subfigure}[b]{0.72\textwidth}
        \centering
        \includegraphics[scale=1]{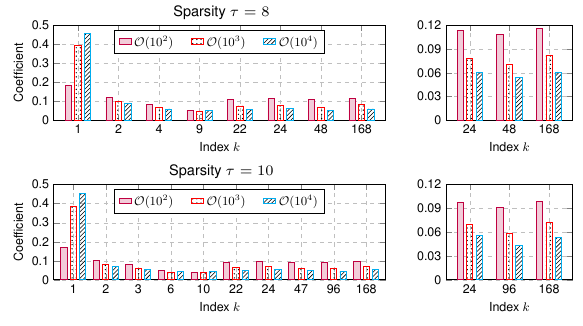}
        \caption{Dominant auto-correlations}
    \end{subfigure}
    \hfill
    \begin{subfigure}[b]{0.27\textwidth}
        \centering
        \includegraphics[scale=1]{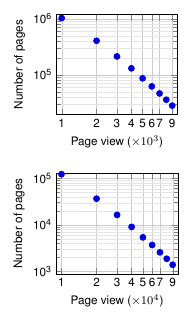}
        \caption{Log-log plot}
    \end{subfigure}

\caption{Seasonality analysis of Wikipedia page view time series with the sparse autoregression model. (a) Auto-correlations on the optimized index set that controlled by the sparsity levels $\tau=8,10$. (b) Log-log plot of distributions of page view data. The top panel shows 9 sub-categories with total number of page views $\mathcal{O}(10^3)$ in January 2024. The bottom panel refers to $\mathcal{O}(10^4)$.}
\label{seasonality}
\end{figure}

\textbf{Experimental Setting}. We conduct extensive experiments to examine the sparse autoregression model on the Wikipedia page view time series dataset. 
The Python implementation is available at \url{https://github.com/xinychen/wikidata}. 
All tests are executed on a workstation equipped with Intel Xeon Silver 4210R CPU, 128GB RAM and four Nvidia Quadro RTX 6000 GPUs. 
The dataset is organized as a three-dimensional array, including Wikipedia page, category, and hourly time step dimensions. Sample page view time series are selected according to three categories of the number of page views in January 2024, i.e., $[10^2,10^3)$, $[10^3,10^4)$, and $[10^4,10^5)$, which are simplified as $\mathcal{O}(10^2)$, $\mathcal{O}(10^3)$, and $\mathcal{O}(10^4)$, respectively. The page numbers in these three categories are 0.78 million, 2.04 million, and 0.20 million, respectively. Although the matrices of three categories have different numbers of rows, one can compute $\boldsymbol{\Phi}_{\gamma}\in\mathbb{R}^{d\times d}$ and $\boldsymbol{\psi}_{\gamma}\in\mathbb{R}^{d}$ in advance, leading to the quadratic objective function in Eq.~\eqref{sparse_autoregression_opt_quad}. We use the MIQP algorithm in CPLEX \cite{cplex_manual} to solve sparse autoregression in Eq.~\eqref{sparse_autoregression_opt_quad} and set the model's sparsity levels as $\tau=8,10$ and the order as $d=168$ (i.e., weekly cycle). The sparse coefficient vectors $\{\boldsymbol{w}_{\gamma}\}_{\gamma\in[3]}$ are visualized in Figure~\ref{seasonality}.

\textbf{Seasonality Results}. As can be seen, under sparsity level $\tau=8$, the support set is optimized as $\Omega=\{1,2,4,9,22,24,48,168\}$ (i.e., cardinality of this set is 8) where indices $\{24,48,168\}$ correspond to the daily, bi-daily, and weekly cycles, respectively. Therefore, the coefficients at index $24$ represent the strength of daily seasonality. Accordingly, we can quantify the bi-daily and weekly seasonality by the coefficients at indices $48$ and $168$, respectively. By contrast, setting the sparsity level as $\tau=10$ leads to the support set $\Omega=\{1,2,3,6,10,22,24,47,96,168\}$, in which indices $\{24,96,168\}$ correspond to the daily, four-day, and weekly cycles, respectively. In these coefficients with the sparsity level $\tau=8$ across three different page categories, one can find that the page view seasonality of frequently-viewed pages is lower than less-viewed pages. In the case of $\tau=10$, one can also see that the time series of frequently-view pages are less seasonal than the less-viewed pages. These findings can explain that the frequently-viewed pages might be venerable to special events and anomalies on digital platforms such as Wikipedia.

\subsection{Prediction Performance}

\begin{figure}[ht!]
    \centering
    \includegraphics[width=1\linewidth]{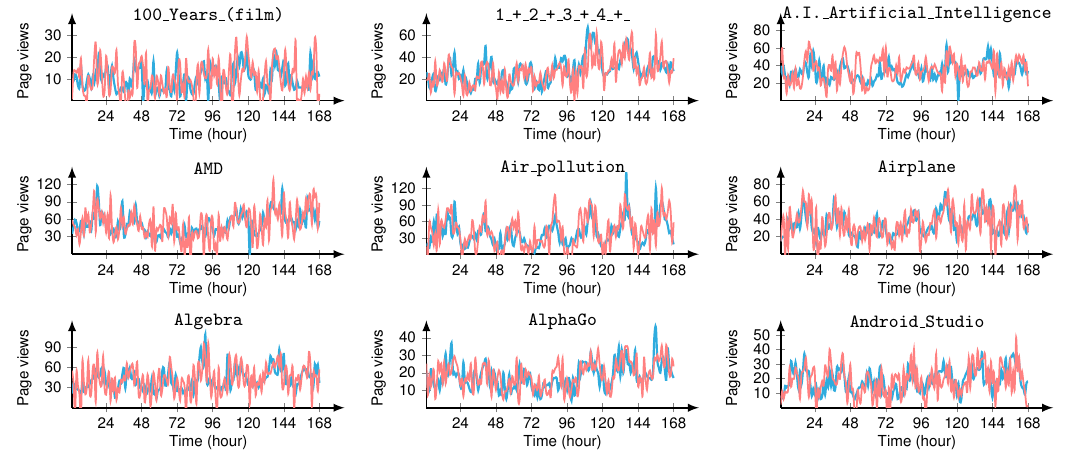}
    \caption{Prediction results of the $\ell_p$-norm autoregression model on selected Wikipedia pages from the $\mathcal{O}(10^4)$ category during the test week. Each panel displays the ground truth hourly page views (cyan) alongside the model's one-step-ahead predictions (red) over a 168-hour horizon. The nine pages span topics in film, mathematics, technology, science, and general knowledge.}
    \label{ts_example}
\end{figure}

To examine the robustness of our proposed framework, we evaluate its predictive performance against standard autoregressive models using Wikipedia page view data, in which the training set is the data from January 1 to January 17, and the data of January 18--24 and January 25--31 are validation set and test set, respectively. The hyperparameters are tuned using the validation set, including Huber threshold $\delta=1$, quantile $\tau=0.3$, and power $p=1/2$. As illustrated in the comparative analysis, while the traditional $\ell_2$-norm loss yields competitive results on lower-magnitude categories ($\mathcal{O}(10^2)$), its performance degrades significantly on high-volume, $\mathcal{O}(10^4)$ categories, manifesting in an RMSE of 1163.08. In contrast, robust estimators such as specifically the Huber and $\ell_p$-norm losses leverage non-Gaussian residual modeling to mitigate the influence of extreme outliers, consistently achieving superior predictive accuracy, see e.g., Figure~\ref{ts_example}. These findings underscore the necessity of selecting appropriate loss functions and domain-aware constraints when modeling real-world, non-stationary time series characterized by heavy-tailed distributions and sparse, zero-inflated observations.

\begin{table}[h]
    \caption{Prediction performance (in MAPE/RMSE) of autoregression models on the Wikpedia data.}
    \centering
    \begin{tabular}{c|c|ccc}
    \toprule
    Loss function & Residual $\varepsilon_t$ & Category $\mathcal{O}(10^2)$ & Category $\mathcal{O}(10^3)$ & Category $\mathcal{O}(10^4)$ \\
    \midrule
    $\ell_2$-norm loss & Gaussian & 1.04/2.39 & 1.56/37.06 & 4.68/1163.08 \\
    $\ell_1$-norm loss & Non-Gaussian & 1.15/2.53 & 1.45/10.91 & 1.52/142.44 \\
    Huber loss & Non-Gaussian & \textbf{1.01}/\textbf{2.31} & 1.32/10.35 & 1.46/139.87 \\
    Quantile loss & Non-Gaussian & 1.17/2.64 & 1.43/10.76 & 1.48/132.14 \\
    $\ell_{p}$-norm loss & Non-Gaussian & 1.02/2.32 & \textbf{1.28}/\textbf{10.03} & \textbf{1.34}/\textbf{129.28} \\
    \bottomrule
    \end{tabular}    \label{prediction_table}
\end{table}

% To account for the zero-inflated nature of the dataset, we constrained the predictions to non-negative values by setting $\hat{x}_{t+1} := \max\{0, \hat{x}_{t+1}\}$. This adjustment improved the performance of the Huber loss model: for the $\mathcal{O}(10^2)$ category, MAPE decreased from 1.01 to 0.84 and RMSE from 2.32 to 2.00. Similarly, for the $\mathcal{O}(10^3)$ category, the errors dropped to a MAPE of 1.10 and an RMSE of 9.14. These results underscore the importance of addressing zero-inflation in the modeling process. Therefore, addressing the challenges of heavy-tailed noises, zero-inflated observations, and high-dimensional settings makes this TailedTS dataset particularly crucial for time series modeling.

\section{Results on Benchmark Datasets}

% How to verify that the residual is heavy-tailed distributed? Prediction performance comparison between heavy-tailed model and Gaussian model.

To verify that the residuals of benchmark time series datasets are heavy-tailed distributed, we examine the log-log histograms of four widely used datasets: UCI electricity load diagrams, SF traffic occupancy, Air quality indices, and Meteorological weather variables. As shown in Figure~\ref{data_histogram_log}, the frequency distributions of these datasets exhibit an approximately linear decay on the log-log scale, which is a hallmark of heavy-tailed behavior. In a Gaussian distribution, the histogram would instead show a rapid, super-exponential drop-off in the tails. The persistence of large values at non-negligible frequencies across all four datasets confirms that extreme observations or outliers occur far more often than Gaussian models would predict, motivating the use of robust, non-Gaussian loss functions for autoregressive prediction on these benchmarks.

To evaluate the practical impact of loss function choice on prediction performance, we apply the five autoregressive models to each of the four benchmark datasets, reporting MAPE and RMSE in Table~\ref{benchmark_prediction_table}. The results consistently demonstrate that non-Gaussian loss functions offer meaningful improvements over the standard $\ell_2$-norm loss across most datasets. On the UCI electricity data, the $\ell_2$-norm, Huber, and $\ell_p$-norm losses all achieve a MAPE of 0.083, compared to 0.095 under the $\ell_2$-norm loss, representing a roughly 13\% relative reduction. Similarly, on the Air quality dataset, the $\ell_1$-norm loss attains the lowest MAPE of 0.124, substantially outperforming the $\ell_2$-norm baseline of 0.156. These gains align with the heavy-tailed structure revealed in Figure~\ref{data_histogram_log}: since the $\ell_2$-norm loss disproportionately penalizes large residuals, it becomes destabilized by the frequent extreme values in these datasets, whereas robust losses such as $\ell_1$ and Huber limit the influence of such outliers during optimization. Taken together, these results underscore the importance of tailoring the choice of loss function to the stochastic properties of the data, and reinforce the value of the TailedTS benchmark in providing a rigorous testbed for evaluating model robustness under non-Gaussian noises.

\begin{table}[h]
    \caption{Prediction performance (in MAPE/RMSE) of autoregression models on the benchmark.}
    \centering
    \begin{tabular}{c|cccc}
    \toprule
    Loss function & UCI electricity data & SF traffic data & Air quality data & Weather data \\
    \midrule
    $\ell_2$-norm loss & 0.095/230.76 & 0.377/0.0168 & 0.156/11.35 & 0.589/\textbf{4.079} \\
    $\ell_1$-norm loss & \textbf{0.083}/\textbf{220.22} & 0.308/0.0179 & \textbf{0.124}/10.64 & 0.450/4.253 \\
    Huber loss & \textbf{0.083}/220.34 & 0.328/0.0175 & 0.133/\textbf{10.63} & 0.451/4.108 \\
    Quantile loss & 0.088/236.87 & \textbf{0.290}/\textbf{0.0167} & 0.141/11.26 & \textbf{0.406}/4.541 \\
    $\ell_{p}$-norm loss & \textbf{0.083}/222.56 & 0.377/0.0168 & 0.140/10.64 & 0.474/4.087 \\
    \bottomrule
    \end{tabular}    \label{benchmark_prediction_table}
\end{table}

\section{Conclusion}

In this work, we introduced the 2024 Wikipedia Hourly Page View Dataset, a massive-scale benchmark specifically designed to challenge the robustness of time series forecasting models. Through our systematic evaluation of various autoregressive baselines, we demonstrated that standard Gaussian-based loss functions which underpin much of the current SOTA are not well-suited to handle the heavy-tailed noise and extreme volatility inherent in real-world digital traffic. Our findings reveal that while robust estimators such as the Huber and $\ell_p$-norm losses offer significant error reduction, the challenges posed by extreme scale ($\mathcal{O}(10^4)$) and pervasive zero-inflation remain a formidable frontier for existing architectures. 

By providing 24.69 billion data points in a high-performance Apache Parquet format, this dataset serves as more than just a repository; it is standardized for evaluating predictive stability under non-Gaussian conditions. We define clear benchmarking tasks that require models to maintain accuracy across four orders of magnitude while navigating the structural constraints of non-negativity. Ultimately, we hope this benchmark catalyzes the development of next-generation, distribution-aware models that are as resilient as they are scalable, bridging the gap between theoretical optimization and heavy-tailed realities of large-scale time series data.

\textbf{Limitations.} 
The proposed benchmark is derived from Wikipedia page view data and does not capture all forms of real-world nonstationarities, such as abrupt regime shifts or adversarial manipulation. As with any domain specific dataset, conclusions drawn from this benchmark may not fully generalize to time series governed by different data generating mechanisms.

\textbf{Broader Impacts.}
The proposed benchmark provides a controlled setting for studying model behavior under extreme variability, enabling the identification of robustness gaps that may be overlooked by standard evaluations and supporting more responsible model development in high‑impact domains.  
At the same time, the dataset is intended for research and evaluation purposes, and insights derived from it should be applied with caution when informing real‑world decisions.

% In this work, ...

% The Wikipedia page view time series dataset is both heavy-tailed and zero-inflated, including massive outliers and irregular periodic patterns. From our preliminary experimental analysis, we have some claims: i) Predicting these time series is challenging because of the nature of non-Gaussian residuals, demanding the reformulation of losses in machine learning models; ii) The Wikipedia dataset is high-dimensional with millions of time series sequences, demanding scalable predictive models; iii) The Wikipedia dataset is zero-inflated with massive zero page views in time series.

% Time-evolving, nonstationary. Prediction with an online machine learning?

% Causal inference...

% \bibliographystyle{abbrvnat} % NeurIPS often uses abbrvnat or unsrtnat
\bibliographystyle{unsrt} % NeurIPS often uses abbrvnat or unsrtnat
\bibliography{references}    % Points to references.bib (do not add .bib extension)

\newpage

\appendix

\begin{center}
\LARGE
  Supplementary Material for \\
  TailedTS: Benchmark Dataset for Heavy-Tailed Time\\ Series Prediction and Periodicity Quantification
\end{center}

\section{Benchmark Datasets}

\begin{figure}[ht!]
\centering
\begin{subfigure}[b]{0.24\textwidth}
\centering
\includegraphics[height=2.65cm]{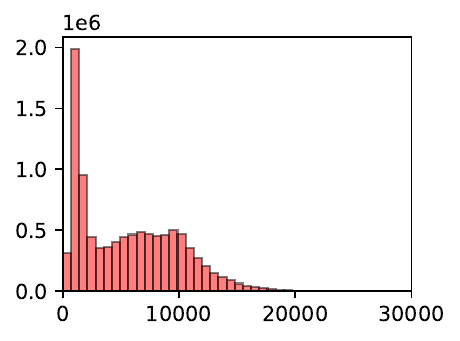}
\caption{M4 data \cite{makridakis2020m4}}
\end{subfigure}
\begin{subfigure}[b]{0.24\textwidth}
\centering
\includegraphics[height=2.65cm]{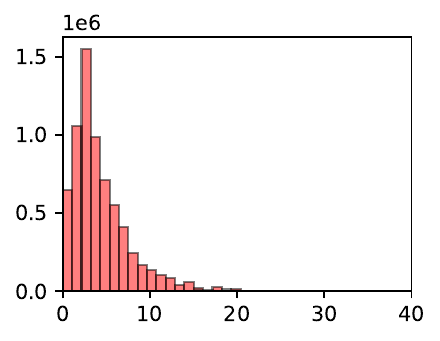}
\caption{M5 retail data \cite{makridakis2022m5}}
\end{subfigure}
\begin{subfigure}[b]{0.24\textwidth}
\centering
\includegraphics[height=2.65cm]{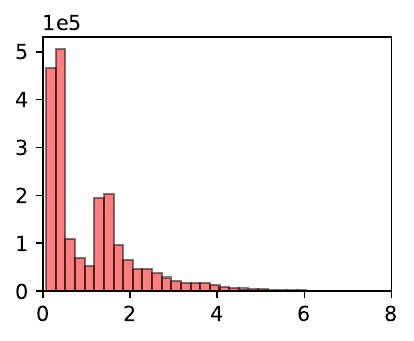}
\caption{Power data \cite{hebrail2012individual}}
\end{subfigure}
\begin{subfigure}[b]{0.24\textwidth}
\centering
\includegraphics[height=2.65cm]{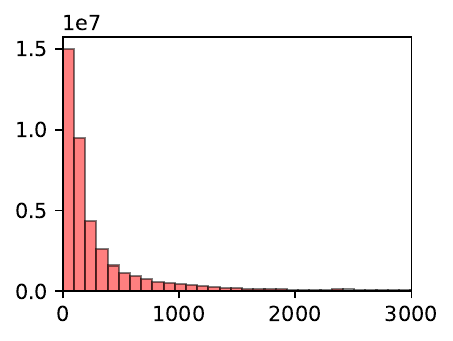}
\caption{UCI electricity data \cite{electricityloaddiagrams20112014_321}}
\end{subfigure}
\begin{subfigure}[b]{0.24\textwidth}
\centering
\includegraphics[height=2.65cm]{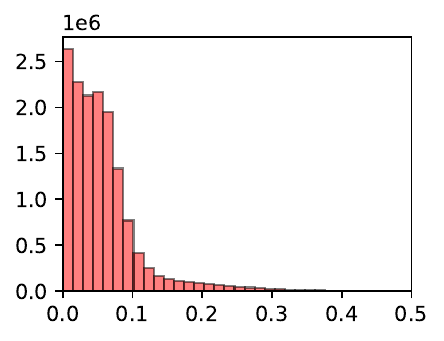}
\caption{SF traffic data \cite{lai2018modeling}}
\end{subfigure}
\begin{subfigure}[b]{0.24\textwidth}
\centering
\includegraphics[height=2.65cm]{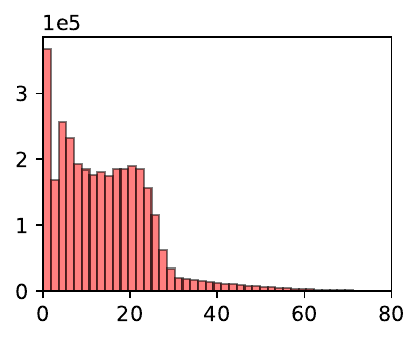}
\caption{Solar energy data \cite{lai2018modeling}}
\end{subfigure}
\begin{subfigure}[b]{0.24\textwidth}
\centering
\includegraphics[height=2.65cm]{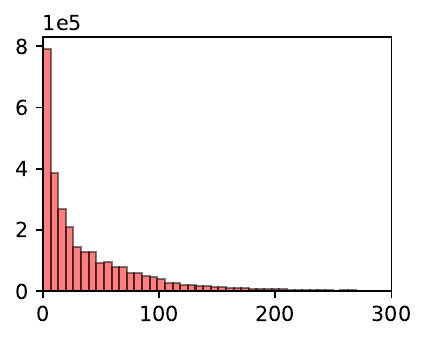}
\caption{Air quality data \cite{godahewa_2020_4656756}}
\end{subfigure}
\begin{subfigure}[b]{0.24\textwidth}
\centering
\includegraphics[height=2.65cm]{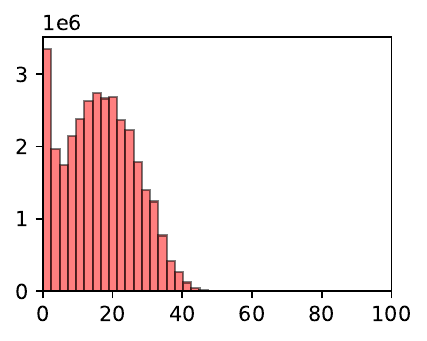}
\caption{Weather data \cite{godahewa_2020_4654822}}
\end{subfigure}
\caption{Empirical distributions of benchmark time series datasets. The histograms illustrate the frequency distribution of values across eight widely used forecasting benchmarks: (a) M4 competition data, (b) M5 retail data, (c) Individual household power consumption, (d) UCI electricity load diagrams, (e) San Francisco (SF) traffic occupancy, (f) Solar energy production, (g) Air quality indices, and (h) Meteorological weather variables.}
\label{data_histogram}
\end{figure}

Figure~\ref{data_histogram} shows the histogram plots of 8 benchmark time series datasets.

\section{Dataset Description}

\subsection{Downloading Source Data Files}

We download the source data files of page view observations from \url{https://dumps.wikimedia.org/other/pageviews/} ({\color{red}license TBD}). The data files of January 2024 is specified in the folder \texttt{2024-01}, i.e., \url{https://dumps.wikimedia.org/other/pageviews/2024/2024-01/}. Since the data is given with each hour, we download 744 data files across 24 hours and different days of January 2024.

\begin{lstlisting}
year = 2024
month = 1

import requests
import os
from datetime import datetime, timedelta

def download_specific_pageviews_files(year, month):
    """
    Download pageviews files
    """
    base_url = f"https://dumps.wikimedia.org/other/pageviews/{year:04d}/{year:04d}-{month:02d}/"
    output_dir = "pageviews_data"
    os.makedirs(output_dir, exist_ok=True)
    
    files_to_download = []
    for day in range(1, 32):
        for hour in range(24):
            filename = f"pageviews-{year:04d}{month:02d}{day:02d}-{hour:02d}0000.gz"
            files_to_download.append(filename)
        
    # Download each file
    for filename in files_to_download:
        file_url = base_url + filename
        local_path = os.path.join(output_dir, filename)
        
        try:
            print(f"Downloading: {filename}")
            response = requests.get(file_url)
            response.raise_for_status()
            
            with open(local_path, 'wb') as f:
                f.write(response.content)
            
            print(f"Saved: {local_path}")
            
        except requests.exceptions.RequestException as e:
            print(f"Failed to download {filename}: {e}")

# Run the download
download_specific_pageviews_files(year, month)
\end{lstlisting}

\subsection{Creating Hourly Page View Time Series}

\textbf{Data Alignment across Hours}. As shown in Figure~\ref{wikipedia_page_dist}(b), the number of Wikipedia pages that have been viewed during each hour is not fixed, varying from 4 million to 8 million. Thus, the first task is Wikipedia page alignment and truncation across different hours of each day.

\begin{lstlisting}
import pandas as pd
import numpy as np
import time

year = 2024
month = 1

for day in range(1, 32):
    start = time.time()
    print('Processing Day:', day)
    df = pd.read_csv(f'pageviews_data/pageviews-{year:04d}{month:02d}{day:02d}-000000.gz', sep=' ', 
                     names = ['domain_code', 'page_title', f'count_views_{day:02d}00', 'total_response_size'], header = None,
                     on_bad_lines = 'skip', engine = 'python')
    df = df.drop(columns = 'total_response_size')
    df = df.fillna({'domain_code': 'NA', 'page_title': 'NA'})
    df = df.groupby(['domain_code', 'page_title'])[f'count_views_{day:02d}00'].sum().reset_index()
    for hour in range(1, 24):
        df_new = pd.read_csv(f'pageviews_data/pageviews-{year:04d}{month:02d}{day:02d}-{hour:02d}0000.gz', sep=' ', 
                              names = ['domain_code', 'page_title', f'count_views_{day:02d}{hour:02d}', 'total_response_size'], header = None,
                              on_bad_lines = 'skip', engine = 'python')
        df_new = df_new.drop(columns = 'total_response_size')
        df_new = df_new.fillna({'domain_code': 'NA', 'page_title': 'NA'})
        df_new = df_new.groupby(['domain_code', 'page_title'])[f'count_views_{day:02d}{hour:02d}'].sum().reset_index()
        df = pd.merge(df, df_new, on = ['domain_code', 'page_title'], how='outer').fillna(0)
    vec = np.sum(df.iloc[:, 2 :].values, axis = 1)
    df_small = df[vec >= 10].reset_index(drop = True)
    df_small.to_csv(f'data-2024{month:02d}{day:02d}.csv.gz', index = False, compression = 'gzip')
    end = time.time()
    print('Running time (s): ', end - start)
    print()
    print()
\end{lstlisting}

\bigskip

\textbf{Data Alignment across Days}. As a result, we have the subset of each day formatted as e.g., \texttt{data-20240101.csv.gz} with a compressed CSV data file for January 1, 2024. The next step is Wikipedia data alignment and truncation across different days of each month. We first need to extract the unique Wikipedia pages across different days.

\begin{lstlisting}
import numpy as np
import pandas as pd

month = 1
day = 1

print('Day:', day)
data = pd.read_csv(f'data-2024{month:02d}{day:02d}.csv.gz', compression='gzip')
df = data[['domain_code', 'page_title']].copy()
df['count_views'] = np.sum(data.iloc[:, 2 :].values, axis = 1)
df = df[df['count_views'] >= 10]
df = df[['domain_code', 'page_title']]
del data

for day in range(2, 32):
    print('Day:', day)
    data_new = pd.read_csv(f'data-2024{month:02d}{day:02d}.csv.gz', compression='gzip')
    df_minimal = data_new[['domain_code', 'page_title']].copy()
    df_minimal['count_views'] = np.sum(data_new.iloc[:, 2 :].values, axis = 1)
    df_minimal = df_minimal[df_minimal['count_views'] >= 10]
    del data_new
    df = pd.merge(df, df_minimal, on = ['domain_code', 'page_title'], how = 'inner')
    df = df[['domain_code', 'page_title']]
df.to_csv(f'data-pages-2024{month:02d}.csv', index = False)
\end{lstlisting}

\bigskip

Then, we use the data file of unique pages in e.g., \texttt{data-pages-202401.csv} to index the original data files across 31 days of January 2024. By doing so, we get the data file \texttt{data-202401.parquet} for January 2024.

\begin{lstlisting}
import pandas as pd
import numpy as np

month = 1
t = 31

page_ind = pd.read_csv(f'data-pages-2024{month:02d}.csv')
page_ind = page_ind.fillna({'domain_code': 'NA', 'page_title': 'NA'})
df1 = page_ind.iloc[: int(1e+6)]
for day in range(1, t + 1):
    print('Processing Day:', day)
    data = pd.read_csv(f'data-5v-2024{month:02d}{day:02d}.csv.gz', compression='gzip')
    data = data.fillna({'domain_code': 'NA', 'page_title': 'NA'})
    df1 = pd.merge(df1, data, on=['domain_code', 'page_title'], how='left')
    del data

df2 = page_ind.iloc[int(1e+6) : int(2e+6)]
for day in range(1, t + 1):
    print('Processing Day:', day)
    data = pd.read_csv(f'data-5v-2024{month:02d}{day:02d}.csv.gz', compression='gzip')
    data = data.fillna({'domain_code': 'NA', 'page_title': 'NA'})
    df2 = pd.merge(df2, data, on=['domain_code', 'page_title'], how='left')
    del data

df3 = page_ind.iloc[int(2e+6) :]
for day in range(1, t + 1):
    print('Processing Day:', day)
    data = pd.read_csv(f'data-5v-2024{month:02d}{day:02d}.csv.gz', compression='gzip')
    data = data.fillna({'domain_code': 'NA', 'page_title': 'NA'})
    df3 = pd.merge(df3, data, on=['domain_code', 'page_title'], how='left')
    del data

df = pd.concat([df1, df2, df3], ignore_index=True)
df.to_parquet(f'data-2024{month:02d}.parquet', compression='zstd')
\end{lstlisting}

\bigskip

\begin{figure}[ht!]
\centering
\includegraphics[width=1\linewidth]{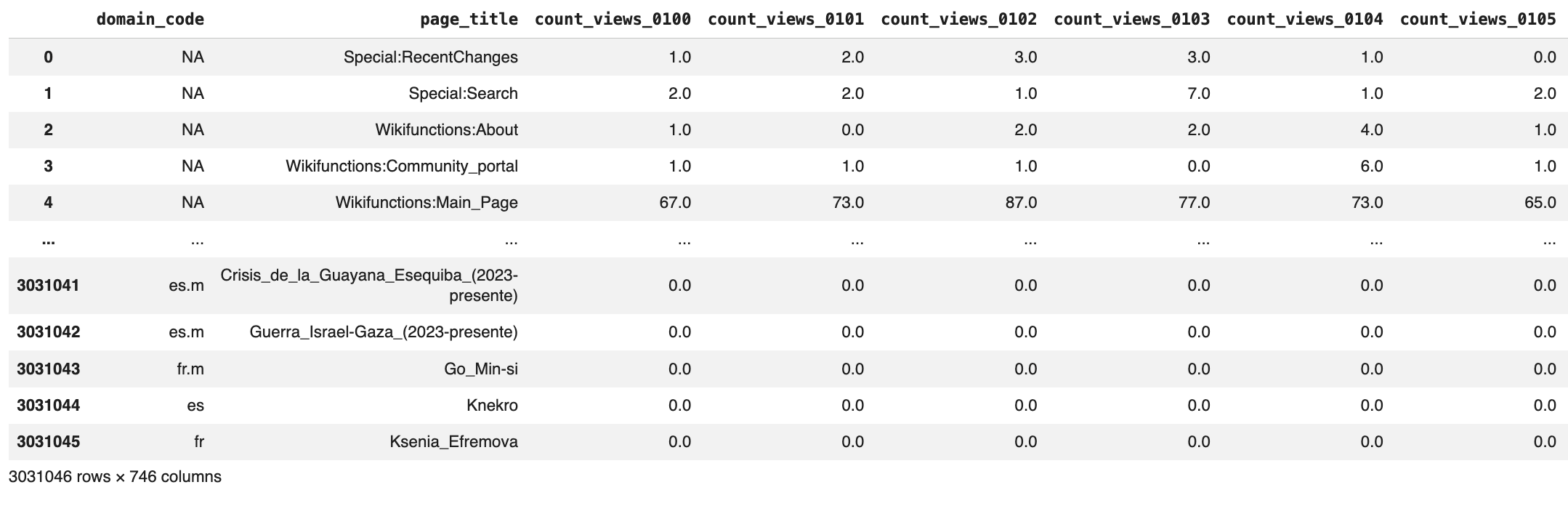}
\caption{Sample of the Wikipedia page view time series dataset (January 2024). The dataset consists of approximately 3.03 million unique page entries (rows) across 744 temporal features (columns). Columns labeled \texttt{count\_views\_MMDD} represent hourly or daily granular observation points for specific wiki domains (e.g., \texttt{es.m} for Spanish mobile, \texttt{fr} for French desktop) and page titles. The matrix illustrates a high degree of sparsity, particularly in non-English domains.}
\label{data202401}
\end{figure}

\begin{figure}[ht!]
\centering
\includegraphics[width=1\linewidth]{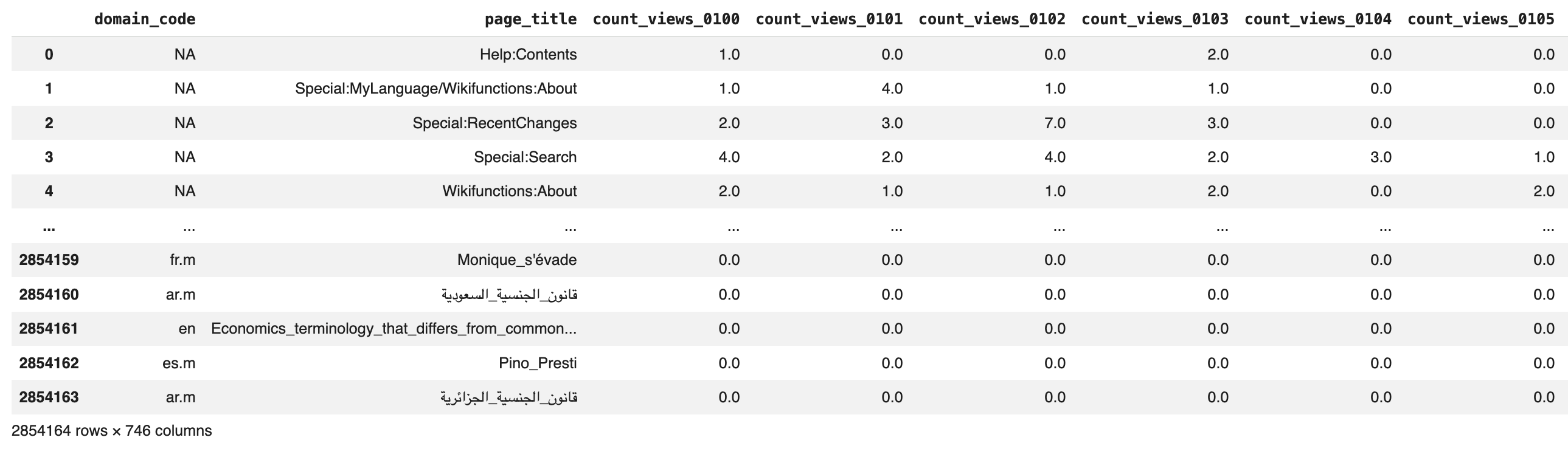}
\caption{Sample of the Wikipedia page view time series dataset (May 2024). The dataset consists of approximately 2.85 million unique page entries (rows) across 744 temporal features (columns).}
\label{data202405}
\end{figure}

Figures~\ref{data202401} and \ref{data202405} show the \texttt{pandas.dataframe} of January 2024 and May 2024, respectively. As can be seen, the data file has the first two columns as \texttt{domain\_code} and \texttt{page\_title}.
\begin{itemize}
\item \texttt{domain\_code}: This field identifies which Wikipedia project and access platform the traffic came from. It is an abbreviated code rather than a full URL.
\begin{itemize}
\item[$\circ$] \textbf{Project Part}: The first part is the language or project code (e.g., \texttt{en} for English Wikipedia, \texttt{de} for German, \texttt{commons} for Wikimedia Commons).
\item[$\circ$] \textbf{Platform Part}: If it ends in \texttt{.m}, it indicates the mobile site. If there is no suffix, it usually refers to the desktop site. In particular, \texttt{en}: English Wikipedia (Desktop); \texttt{en.m}: English Wikipedia (Mobile); \texttt{es.m.w}: Spanish Wiktionary (Mobile).
\end{itemize}
\end{itemize}

Table~\ref{data_file_table} summarizes the Wikipedia hourly page view time series dataset for 2024, detailing monthly statistics for data volume, storage size, and page coverage. Throughout 2024, the dataset consistently captured traffic for approximately 2.6 to 3.0 million unique Wikipedia pages per month, with January and February seeing the highest concentration of active pages. This results in a massive computational scale of roughly 1.92 to 2.25 billion discrete data points monthly, totaling approximately 24.69 billion observations for the full year. Despite the high-dimensional nature of the hourly time series, the use of the Apache Parquet columnar format ensures high storage efficiency; monthly files are compressed to a manageable footprint ranging from 1.22 GB to 1.49 GB, demonstrating the technical advantages of binary storage formats for large-scale data science and research applications.

\begin{table}[h]
    \caption{Summary of Wikipedia page view datasets for 2024. Each monthly subset captures traffic for approximately 3 million unique pages, resulting in 2 billion discrete data points per month. By utilizing the Parquet columnar format, the high-dimensional data (originally exceeding 1.9 billion points) is compressed to a manageable footprint of approximately 1.3\,GB per file/month. This highlights the efficiency of binary storage formats for large-scale time series data.}
    \centering
    \begin{tabular}{c|ccccc}
    \toprule
    Time & Data file name & Size & Wiki pages & Data points & Zero points \\
    \midrule
    January 2024 & \texttt{data-202401.parquet} & 1.49\,GB & 3,031,046 & 2.25\,B & 24.38\% \\
    February 2024 & \texttt{data-202402.parquet} & 1.43\,GB & 3,032,833 & 2.11\,B & 25.01\% \\
    March 2024 & \texttt{data-202403.parquet} & 1.44\,GB & 2,918,686 & 2.17\,B & 24.53\% \\
    April 2024 & \texttt{data-202404.parquet} & 1.39\,GB & 2,951,902 & 2.13\,B & 24.69\% \\
    May 2024 & \texttt{data-202405.parquet} & 1.34\,GB & 2,854,164 & 2.12\,B & 24.58\% \\
    June 2024 & \texttt{data-202406.parquet} & 1.22\,GB & 2,671,128 & 1.92\,B & 24.52\% \\
    July 2024 & \texttt{data-202407.parquet} & 1.27\,GB & 2,624,815 & 1.95\,B & 24.44\% \\
    August 2024 & \texttt{data-202408.parquet} & 1.28\,GB & 2,659,168 & 1.98\,B & 24.30\% \\
    September 2024 & \texttt{data-202409.parquet} & 1.26\,GB & 2,706,142 & 1.95\,B & 24.59\% \\
    October 2024 & \texttt{data-202410.parquet} & 1.30\,GB & 2,751,244 & 2.05\,B & 24.54\% \\
    November 2024 & \texttt{data-202411.parquet} & 1.25\,GB & 2,743,321 & 1.98\,B & 24.64\% \\
    December 2024 & \texttt{data-202412.parquet} & 1.27\,GB & 2,625,481 & 1.95\,B & 24.10\% \\
    \bottomrule
    \end{tabular}    \label{data_file_table}
\end{table}

\begin{lstlisting}
import pandas as pd

month = 1
df = pd.read_parquet(f'data-2024{month:02d}.parquet')
\end{lstlisting}

\section{Optimization over Non-Gaussian Losses}

In what follows, we give the detailed optimization scheme for different non-Gaussian losses of autoregressive prediction.

\subsection{Huber Loss}\label{huber_opt_section}

\subsubsection{Quadratic Programming (QP)}

The optimization of Huber loss can be written as a QP problem \cite{boyd2004convex} such that
\begin{equation}
\begin{aligned}
\min_{\boldsymbol{w},\boldsymbol{\alpha},\boldsymbol{\beta}}\quad&\|\boldsymbol{\beta}\|_2^2+2\delta\|\boldsymbol{\alpha}\|_1 \\
\text{s.t.}\quad&\varepsilon_{t}=x_{t}-\sum_{k\in[d]}w_{k}x_{t-k}, &\forall t\in[d+1,T], \\
&-\alpha_{t}\leq\varepsilon_{t}-\beta_{t}\leq\alpha_t,\quad\alpha_t\geq0, &\forall t\in[d+1,T],
\end{aligned}
\end{equation}
where $\boldsymbol{\alpha},\boldsymbol{\beta}\in\mathbb{R}^{T-d}$ are slack vectors.

\subsubsection{IRLS Method}

The optimization of Huber loss can also be formulated with the IRLS method such that
\begin{equation}
\begin{aligned}
\boldsymbol{w}^{(i+1)}:&=\arg\min_{\boldsymbol{w}}\quad\sum_{t\in[d+1,T]}\vartheta_{t}^{(i)}\varepsilon_{t}^2 \quad\quad\text{s.t.}\quad\varepsilon_{t}=x_{t}-\sum_{k\in[d]}w_{k}x_{t-k},\quad\forall t\in[d+1,T], \\
&=\bigl(\boldsymbol{A}^\top\operatorname{diag}(\boldsymbol{\vartheta})\boldsymbol{A}\bigr)^{-1}\boldsymbol{A}^\top\operatorname{diag}(\boldsymbol{\vartheta})\boldsymbol{y}, \\
\vartheta_{t}^{(i+1)}:&=\min\{1,\delta/|\varepsilon_{t}|\}, \quad\quad\text{s.t.}\quad\varepsilon_{t}=x_{t}-\sum_{k\in[d]}w_{k}^{(i+1)}x_{t-k},\quad\forall t\in[d+1,T],
\end{aligned}
\end{equation}
with
\begin{equation}
\begin{cases}
\boldsymbol{y}\triangleq\{x_{t}\}_{t\in[d+1,T]}\in\mathbb{R}^{T-d}, \\
\boldsymbol{\vartheta}\triangleq\{\vartheta_{t}\}_{t\in[d+1,T]}\in\mathbb{R}^{T-d}, \\
\boldsymbol{y}^{(k)}\triangleq\{x_{t}\}_{t\in[d+1-k,T-k]}\in\mathbb{R}^{T-d},\quad\forall k\in[d], \\
\boldsymbol{A}\triangleq\{\boldsymbol{y}^{(k)}\}_{k\in[d]}\in\mathbb{R}^{(T-d)\times d}. \\
\end{cases}
\end{equation}

In this experiment, we validate the Huber threshold as $\delta=1$ and maximum iteration $20$ in IRLS.

% We provide the comparison between QP and IRLS in Table.
% \begin{itemize}
% \item Different number of iterations
% \item Computational time (cvxpy vs. numpy) on different categories.
% \end{itemize}

\subsection{Quantile Loss}\label{quantile_opt_section}

The Linear Programming (LP) problem of quantile loss can be formulated as follows,
\begin{equation}
\begin{aligned}
\min_{\boldsymbol{w},\{\varepsilon_{t}^{+},\varepsilon_{t}^{-}\}_{t\in[d+1,T]}}\quad&\sum_{t\in[d+1,T]}\tau\varepsilon_{t}^{+}+(1-\tau)\varepsilon_{t}^{-} \\
\text{s.t.}\quad&\varepsilon_{t}^{+}-\varepsilon_{t}^{-}=x_t-\sum_{k\in[d]}w_{k}x_{t-k},\quad\varepsilon_t^{+},\varepsilon_{t}^{-}\geq0,\quad\forall t\in[d+1,T],
\end{aligned}
\end{equation}
where $\varepsilon_{t}^{+}$ and $\varepsilon_{t}^{-}$ are over-estimated and under-estimated residual vectors, respectively. 

Suppose quantile $\tau=1/2$, we have the same penalty for under-estimate and over-estimate, referring to the $\ell_1$-norm losses. In particular, the LP problem of $\ell_1$-norm loss can be written as follows,
\begin{equation}
\begin{aligned}
\min_{\boldsymbol{w},\boldsymbol{\alpha}}\quad&\sum_{t\in[d+1,T]}\alpha_{t} \\
\text{s.t.}\quad&-\alpha_{t}\leq x_t-\sum_{k\in[d]}w_{k}x_{t-k}\leq\alpha_t,&\forall t\in[d+1,T], \\
&\alpha_t\geq0,&\forall t\in[d+1,T],
\end{aligned}
\end{equation}
where $\boldsymbol{\alpha}\in\mathbb{R}^{T-d}$ is the bound vector.

\subsection{$\ell_p$-Norm Loss}

The optimization of $\ell_p$-norm loss can be formulated with the IRLS method such that
\begin{equation}
\begin{aligned}
\boldsymbol{w}^{(i+1)}:&=\bigl(\boldsymbol{A}^\top\operatorname{diag}(\boldsymbol{\vartheta})\boldsymbol{A}\bigr)^{-1}\boldsymbol{A}^\top\operatorname{diag}(\boldsymbol{\vartheta})\boldsymbol{y}, \\
\varepsilon_{t}:&=x_{t}-\sum_{k\in[d]}w_{k}^{(i+1)}x_{t-k}, &\forall t\in[d+1,T], \\
\vartheta_{t}^{(i+1)}:&=(\varepsilon_t^2+\epsilon)^{(p-2)/2}, &\forall t\in[d+1,T], \\
\end{aligned}
\end{equation}
where $\epsilon>0$ is the smoothing parameter. We set $p=1/2$ in our experiment. The dynamic smoothing parameter is initialized as $\epsilon_0=100$ and updated by $\epsilon^{(i+1)}:=0.95\epsilon^{(i)}$ in IRLS. For comparison, we also consider $p=1/3$ and $2/3$ in our experiment, see Table~\ref{Lp_pred_table}.

\begin{table}[h]
    \caption{Prediction performance (in MAPE/RMSE) of the autoregression model with $\ell_p$-norm losses.}
    \centering
    \begin{tabular}{c|c|ccc}
    \toprule
    Loss function & Residual $\varepsilon_t$ & Category $\mathcal{O}(10^2)$ & Category $\mathcal{O}(10^3)$ & Category $\mathcal{O}(10^4)$ \\
    \midrule
    $\ell_{2/3}$-norm loss & Non-Gaussian & -/- & 1.29/10.09 & 1.34/129.44 \\
    $\ell_{1/2}$-norm loss & Non-Gaussian & 1.02/2.32 & 1.28/10.03 & 1.34/129.28 \\
    $\ell_{1/3}$-norm loss & Non-Gaussian & 1.02/2.31 & 1.28/10.00 & 1.35/129.97 \\
    \bottomrule
    \end{tabular}    \label{Lp_pred_table}
\end{table}

% \section{Model Setting \& Validation}

%%%%%%%%%%%%%%%%%%%%%%%%%%%%%%%%%%%%%%%%%%%%%%%%%%%%%%%%%%%%

% \newpage

% \input{checklist.tex}

\end{document}